# Fast Calculation of Entropy with Zhang's Estimator

*Antoni Lozano[1]*
*Bernardino Casas[2]*
*Chris Bentz[3]*
*Ramon Ferrer-i-Cancho[2,\*]*

(1) COMBGRAF Research Group. Departament de Ciències de la Computació. Universitat Politècnica de Catalunya. Barcelona, Catalonia, Spain.
(2) Complexity and Quantitative Linguistics Lab. LARCA Research Group. Departament de Ciències de la Computació. Universitat Politècnica de Catalunya. Barcelona, Catalonia, Spain.
(3) Department of General Linguistics, University of Tübingen, Tübingen, Germany.
(\*) Corresponding author, rferrericancho@cs.upc.edu

**Abstract.** Entropy is a fundamental property of a repertoire. Here, we present an efficient algorithm to estimate the entropy of types with the help of Zhang's estimator. The algorithm takes advantage of the fact that the number of different frequencies in a text is in general much smaller than the number of types. We justify the convenience of the algorithm by means of an analysis of the statistical properties of texts from more than 1000 languages. Our work opens up various possibilities for future research.

## 1. Introduction

In quantitative linguistics, the Shannon entropy of types is a fundamental property of a repertoire. From a Zipfian perspective, vocabularies are shaped by a tension between unification and diversification forces (Zipf 1949). The entropy of types is a measure of the degree of diversification of word use or, equivalently, a measure of lexical diversity: it takes a value of 0 when only one type is used, while it takes its maximum value when all types are equally likely (Ferrer-i-Cancho 2005; Bentz et al. 2015). Theoretically, entropy is a better measure of vocabulary size than the raw number of different types: the entropy of types measures the effective size of the vocabulary, which is related to the concept of typical set in information theory (Ferrer-i-Cancho 2014). In practice, the problem is that the true number of types is unknown but it is possible to estimate entropy by taking into account that not all types have been observed (Grabchak et al 2013). For similar reasons, entropy is also used as an index of the diversity of species in biology (Chao, Shen 2003, Jost 2006).

From an information theoretic perspective, the entropy of words yields a lower bound to the mean length of words assuming uniquely decipherable coding (Cover, Thomas 2006). Besides, word entropy minimization is a core assumption of information theoretic models of Zipf's law for word frequencies (Ferrer-i-Cancho 2005, Ferrer-i-Cacho, Solé 2003). From a psychological perspective, entropy is hypothesized to be a measure of cognitive cost (Ferrer-i-Cancho 2014).

The Shannon entropy of a text with a vocabulary of *V* types is defined as (Shannon 1951)

$$H = \sum_{i=1}^{V} p_i \log p_i, \qquad (1)$$

where $p_i$ is the probability of the *i*-th type.

If $f_i$ is the absolute frequency of the *i*-th type in a text of *T* tokens, the probability of the *i*-th type can be estimated by the relative frequency of the type as

$$\hat{p}_i = \frac{f_i}{T} \qquad (2)$$

with

$$T = \sum_{i=1}^{V} f_{i.}. \qquad (3)$$

The entropy of a vocabulary can be estimated naively by replacing the probabilities in Eq. 1 with the relative frequencies in Eq. (2). This is the so-called plugin or maximum likelihood entropy estimator that is known to underestimate the true entropy in practical applications (Hausser, Strimmer 2009). Here we focus on an alternative method that reduces the original downwards bias of the naïve estimator: Zhang's entropy estimator (Zhang 2012). Our goal is to provide an efficient algorithm for that estimator.

In the design of an entropy estimator there is often a trade-off between its computational cost and the bias of the estimator. The NSB estimator, which is considered one of the best entropy estimators, is computationally expensive (Hausser, Strimmer 2009), and for this reason had to be discarded in Vu et al's (2007) study.

In general, we consider algorithms to estimate entropy whose input consists of the absolute frequencies of the *V* types. Those algorithms use at least $\theta(V)$ memory for the input frequency of each type. We will refer to the extra memory used by these algorithms as additional memory. It is easy to see that the plugin estimator and the Chao-Shen estimator run in time $\theta(V)$ with $O(1)$ additional memory. The time cost of Zhang's estimator has not been analyzed in detail to our knowledge. For computational purposes, the recommended definetion of Zhang's estimator is (Grabchak et al. 2013)

$$H_z = \sum_{i=1}^{V} \hat{p}_i \sum_{v=1}^{T-f_i} \frac{1}{v} \prod_{j=0}^{v-1} \left(1 + \frac{1-f_i}{T-1-j}\right), \qquad (4)$$

where *V* is hereafter the number of types in a text. Noting that $1 \leq i \leq V$ and $1 \leq v$, $j \leq T$ it is easy to derive an algorithm that runs in $O(VT^2)$ time. Here, we present an algorithm that estimates entropy in $O(WT)$ time, where $W$ is the number of occupied frequencies. Suppose that $n(f)$ is the number of word types that have frequency $f$ in the sample; $n(1),\ldots,n(f),\ldots n(T)$ defines the frequency spectrum of the sample (Tuldava 1996). By definition

$$V = \sum_{f=1}^{T} n(f). \tag{5}$$

$W$ is the number of values of $f$ such that $n(f) > 0$.

The remainder of the article is organized as follows. Section 2 presents a couple of algorithms to estimate entropy: one that runs in $\theta(VT)$ time and uses $O(1)$ additional memory and another that runs in $O(WT)$ time and uses $\theta(f_{max})$ additional memory, where $f_{max}$ is the maximum $f$ such that $n(f)>0$. The second algorithm is potentially faster as $W \leq V$. Section 3 investigates some statistical properties of real texts that are needed to justify the convenience of the $O(WT)$ time algorithm. On the one hand, we study the ratio $W/V$ in a couple of parallel corpora, concluding that the $O(WT)$ algorithm is at least 5 times faster than the $\theta(VT)$ algorithm when used to estimate word entropies in real texts. On the other hand, we study the ratio $f_{max}/V$, concluding that the additional memory required by the $O(WT)$ time algorithm can be neglected compared to $V$, the cost of storing the table of type frequencies. Section 4 discusses the results.

## 2. A Faster Algorithm

In general, entropy estimation algorithms take the $V$ frequencies of every type, i.e. $f_1, \ldots, f_i,\ldots,f_V$, as the input and return an entropy estimate (Hausser, Strimmer 2009, Zhang 2012). Here we consider the particular case of algorithms based on Zhang's estimator.

It is convenient to define Zhang's estimator equivalently as

$$H_z = \frac{1}{T}\sum_{i=1}^{V} f_i Q(f_i), \tag{6}$$

where

$$Q(f) = \sum_{v=1}^{T-f} \frac{1}{v} \prod_{j=0}^{v-1} \left(1 + \frac{1-f}{T-1-j}\right). \tag{7}$$

This decomposition leads to Algorithm A:

1. Set $K$ to 0
2. For each $i$ such that $1 \leq i \leq V$ do
    a. Calculate $Q(f_i)$
    b. Sum $f_i Q(f_i)$ to $K$
3. Return $K/T$

It is easy to see that the running time of Algorithm A is $O(V \, time(Q))$ where $time(Q)$ is the worst running time of the calculation of $Q$ (Step 2.a. of Algorithm A) for a given type. Since $1 \leq v, j \leq T$, an algorithm for the calculation of $Q(f)$ following Eq. 7 naively runs in $O(T^2)$ time and then the running time of Algorithm A becomes $O(VT^2)$. The space cost is given by the size of the input, i.e. $\theta(V)$.

We will show that the running time of the calculation of $Q(f)$ can be reduced to $O(T)$. To see it, it is convenient to define $Q(f)$ as

$$Q(f) = \sum_{v=1}^{T-f} \frac{1}{v} R(v, f), \tag{8}$$

where

$$R(v, f) = \prod_{j=0}^{v-1} \left(1 + \frac{1-f}{T-1-j}\right). \tag{9}$$

A close observation of Eq. 9 allows one to realize that $R(v,f)$ can be defined recursively, i.e. $R(v,f) = 1$ if $v = 0$ and

$$R(v, f) = \left(1 + \frac{1-f}{T-v}\right) R(v-1, f) \tag{10}$$

if $v > 0$. The recursive definition of $R(v,f)$ allows one to calculate $Q(f)$ faster with Algorithm B:

1. Set $Q$ to 0 and $R$ to 1
2. For $v = 1$ to $T - f$ do
   a. Multiply $R$ by $\left(1 + \frac{1-f}{T-v}\right)$
   b. Sum $R/v$ to $Q$
3. Return $Q$

Algorithm B performs of the order of $T - f + 1$ operations (1 for Step 1 and $T - f$ for Step 2). Thus it runs in time $\theta(T-f)$.

The kind of time optimization that we have performed to calculate $Q(f)$ in Algorithm B, namely the addition of extra local variables to recycle computations from previous iterations, is reminiscent of the one that was applied in an efficient algorithm for the moving average type-token ratio (Covington, McFall 2010).

We obtain Algorithm A' by integrating Algorithm B into Algorithm A. It turns out that the number of operations performed by Algorithm A' is of the order of

$$T(A') = \sum_{i=1}^{V}(T - f_i + 1) = V(T+1) - \sum_{i=1}^{V} f_i. \tag{11}$$

Recalling the definition of $T$ in Eq. 3 one obtains finally

$$T(A') = T(V - 1) + V. \qquad (12)$$

Since $V \leq T$, Algorithm A' estimates entropy in time $\theta(VT)$, which is a significant improvement with respect to the naïve Algorithm A that runs in $O(VT^2)$. Algorithm A' also needs $\theta(V)$ space.

The running time of the entropy estimation can be reduced further. To see it, it is convenient to restate Eq. 6 equivalently as

$$H_z = \frac{1}{T} \sum_{f=1}^{f_{max}} n(f) f Q(f). \qquad (13)$$

Notice that one only has to calculate $Q(f)$ when $n(f) > 0$. This formulation leads to Algorithm C:

1. Calculate $f_{max}$ and the frequency spectrum $n(1),\ldots,n(f),\ldots n(f_{max})$ from $f_1$, $\ldots, f_i, \ldots, f_V$
2. Set $K$ to 0
3. For $f = 1$ to $f_{max}$ do
   a. If $n(f) > 0$ then
      i. Calculate $Q(f)$ with Algorithm B
      ii. Sum $n(f) f Q(f)$ to $K$
4. Return $K/T$

Algorithm C calculates $Q(f)$ only for $W$ values of $f$. Assuming the worst case for the calculation of $Q(f)$ with Algorithm B, namely a running time of $O(T)$, one sees that the running time of Algorithm C is $O(WT)$ (the cost of the 1st step, the calculation of $n(f)$, is $O(T)$). This suggests that Algorithm C is faster than Algorithm A' (which runs in $\theta(VT)$ time) because $W \leq V$, with equality if and only if $n(f) \in \{0,1\}$. To see it, notice that $W$ can be defined as

$$W = \sum_{f=1}^{T} b(f), \qquad (14)$$

where $b(f)$ is a binary variable that indicates if $n(f) > 0$ ($b(f) = 1$ if $n(f) > 0$ and $b(f) = 0$ if $n(f) = 0$). Noting that $b(f) \leq n(f)$ and recalling the definition of $V$ in Eq. 5, $W \leq V$ follows easily, with equality if and only if $b(f) = n(f)$, i.e. $n(f) \in \{0,1\}$ for all $f$. The case $n(f) \in \{0,1\}$ never happens in a sufficiently large real text: it only happens for high frequencies as noted by Balasubrahmanyan, Naranan (1996).

Stronger evidence for the superiority of Algorithm C in respect of time efficiency can be obtained easily. On the one hand, the bulk of the time cost of Algorithm C is determined by step 3 and is of the order of

$$T(C) = \sum_{f=1}^{f_{max}} b(f)(T - f + 1). \qquad (15)$$

Step 1 can be carried out in $\theta(\max(V, f_{max}))$ time if one uses a table of size $f_{max}$ to store $n(1),\ldots,n(f),\ldots n(f_{max})$. Recalling the definition of $V$ in Eq. 5, it is easy to show that $T(C) \geq V$ since $f \leq T$.

On the other hand, the cost of Algorithm A' given in Eq. 11 can be expressed equivalently as

$$T(A') = \sum_{f=1}^{f_{max}} n(f)(T - f + 1). \tag{16}$$

Thus, the number of elementary operations saved by Algorithm C (excluding Step 1) is of the order of

$$T(A') - T(C) = \sum_{f=1}^{f_{max}} (n(f) - b(f))(T - f + 1). \tag{17}$$

It is easy to see that $T(A')-T(C) \geq 0$ since $b(f) \leq n(f)$ and $f \leq T$.

The cost of Algorithm C can be calculated with more precision. Expressing Eq. 15 as

$$T(C) = (T+1) \sum_{f=1}^{f_{max}} b(f) - \sum_{f=1}^{f_{max}} b(f)f \tag{18}$$

and recalling the definition of $W$ in Eq. 14, one obtains finally

$$T(C) = W(T+1) - \sum_{f=1}^{f_{max}} b(f)f. \tag{19}$$

Combining Eq. 12 and Eq. 19, the cost saved by Algorithm C with respect to A' becomes

$$T(A') - T(C) = (V - W)(T+1) - T + \sum_{f=1}^{f_{max}} b(f)f. \tag{20}$$

This shows that the extra time cost of calculating $n(f)$ in Step 1 of Algorithm C (which has cost $\theta(\max(V,f_{max})) \subseteq O(T)$) is balanced by a time saving in the remainder of the algorithm provided that $V - W$ is sufficiently large. In the next section, we will see that $V$ is indeed much larger than $W$ in real texts.

Table 1
Time and space cost of the algorithms

| Algorithm | Time | Space |
| --- | --- | --- |
| A | $O(VT^2)$ | $\theta(V)$ |
| A' | $\theta(VT)$ | $\theta(V)$ |
| C | $O(WT)$ | $\theta(\max(V, f_{max}))$ |

$T$ is the number of tokens, $V$ is the number of types, $W$ is the number of different frequencies and $f_{max}$ is the largest frequency where $n(f) > 0$

Algorithm C needs $O(f_{max})$ additional memory if the values $n(1),\ldots,n(f),\ldots n(f_{max})$ are stored in a simple table and thus the overall memory cost of Algorithm C is $\theta(\max(V, f_{max}))$. In contrast, Algorithm A' needs only additional $O(1)$ memory and its overall memory cost is $\theta(V)$. It is crucial to know how big the cost of storing that table is compared to the memory cost of storing the table of type frequencies, that is $\theta(V)$. Table 1 summarizes the time and memory cost of each of the algorithms discussed so far.

In general, further space can be saved if the input is defined differently. If the input is defined by $n(1),\ldots,n(f),\ldots n(f_{max})$, then the space cost of Algorithm C reduces to $\theta(f_{max})$. If the input is defined by a $W \times 2$ matrix whose columns are $f$ and $n(f)$ and whose rows contain only the values of $f$ where $n(f) > 0$, then the space cost of that algorithm becomes $\theta(W)$. These kinds of improvements and their implications for other algorithms are left for future work.

Table 2
Summary of the statistical properties of the texts of the UDHR and the PBC.

|  |  | UDHR | PBC |
|---|---|---|---|
| $T$ | Min | 105 | 2836 |
|  | Mean | 1801.5 | 290392 |
|  | Standard deviation | 536.7 | 215641 |
|  | Max | 4010 | 1257218 |
| $W/V$ | Min | 0.016 | 0.0014 |
|  | Mean | 0.057 | 0.037 |
|  | Standard deviation | 0.027 | 0.031 |
|  | Max | 0.17 | 0.21 |
| $f_{max}/V$ | Min | 0.035 | 0.015 |
|  | Mean | 0.25 | 1.9 |
|  | Standard deviation | 0.19 | 2.4 |
|  | Max | 1.3 | 33 |

$T$ is the length in tokens, $W/V$ is the ratio between the number of different frequencies and the number of types, $f_{max}/V$ is the ratio between maximum frequency and the text length. The statistics on $T$ were rounded to leave only one decimal digit. The statistics on $W/V$ and $f_{max}/V$ were rounded to leave only two significant digits.

## 3. Statistical Analyses

### 3.1. The Datasets

We use the Parallel Bible Corpus (PBC; Mayer, Cysouw 2014) and the Universal Declaration of Human Rights (UDHR) (http://www.unicode.org/udhr/) to obtain the frequency of word types across languages.

  The version of the PBC we use here comprises 1491 texts that have been assigned 1118 unique ISO 639-3 codes, i.e. unique languages. Some languages are represented by several different translations. The PBC text files are semi-automatically processed to delimit word tokens from punctuation marks by white spaces. Note that this is a non-trivial task for some characters such as apostrophes and hyphens. Depending on the script used to write a language, they have to be either interpreted as part of word tokens or as punctuation. Based on the decisions made in the PBC, word tokens are here defined as strings of alphanumeric characters delimited by white spaces.

  The UDHR comprises more than 400 texts with unique ISO codes. However, only 376 of these are fully converted into Unicode. UDHR files do not come with manually checked white spaces between word tokens and punctuation, and hence can bear more noise. We created frequency lists by splitting Unicode strings according to non-word characters, i.e. punctuation and space symbols. For some of the most widespread writing systems (e.g. Latin, Cyrillic, Devanagari and Arabic) the resulting lists of word types were checked by native speakers for misclassified items. In principle, this matches the method described for the PBC. Some of the UDHR texts had to be excluded due to incompleteness, or due to a script that does not support splitting word tokens by white space characters (e.g. Chinese or Mon-Khmer). This yields a sample of 356 UDHR texts with 333 unique ISO codes.

  A summary of statistical properties of the lengths of the texts in each collection is provided in Table 2. The minimum *T* in the PBC (i.e. 2836 tokens) is too low for the whole Bible. It comes from a version of the Bible in Baruya (a language of Papua New Guinea) that has only a few verses translated.

### 3.2. Results

Table 2 summarizes the statistical properties of *W/V* in real texts. Interestingly, *W/V*≈0.057 on average for the UDHR and *W/V*≈0.037 on average for the PBC. This allows one to conclude that Algorithm C is at least 17 times faster than Algorithm A' on average (approximately). Figure 1 suggests a tendency of *W* to decrease as a function of *V* in the UDHR and the PBC. Such a tendency is supported by a Kendall τ correlation test: τ = -0.35 and p-value < $10^{-20}$ for the UDHR, τ = -0.084 for the PBC and p-value < $10^{-5}$. Our findings indicate that the number of main iterations of Algorithm C will tend to decrease as *V* increases. The fact that *V* is much larger than *W* in general (Fig 1) also indicates that the extra cost of Step 1 in Algorithm C can be neglected.

Table 2 also summarizes the statistical properties of $f_{max}/V$ in real texts. Interestingly, $f_{max}/V \approx 0.25$ on average for the UDHR and $f_{max}/V \approx 1.9$ on average for the PBC. These findings indicate that the extra memory cost of Algorithm C is easy to tolerate in general. For the UDHR this is obvious because $f_{max}/V$ does not exceed 1. Concerning the PBC, $f_{max}$ is about two times $V$, but it is possible to store the input as a $W \times 2$ matrix employing $\theta(W)$ space as explained in Section 2 and we have already shown that $W$ is on average 17 times smaller than $V$.

Figure 1 suggests a tendency of $f_{max}$ to decrease as a function of $T$ in the UDHR and the PBC. Such a tendency is supported by a Kendall $\tau$ correlation test: $\tau = -0.24$ and p-value $< 10^{-10}$ for the UDHR and $\tau = -0.073$ and p-value $< 10^{-4}$ for the PBC.

## 4. Discussion

We have provided an efficient algorithm to estimate entropy with less error than the naïve entropy estimator. Our algorithm takes advantage of the fact that $W$ is much smaller than $V$ to save computation time.

Our work opens up new possibilities for future research. First, our algorithm allows us to estimate entropy with Zhang's method in large corpora. Second, our investigation of the relationship between $W$ and $V$ or between $f_{max}$ and $V$ can be seen as emerging topics for research in quantitative linguistics. We have shown that $W$ and $f_{max}$ tend to decrease as $V$ increases, especially for the UDHR. Figure 1 suggests that $W$ decreases linearly with $V$ for the UDHR but the actual functional dependency between $W$ and $V$ should be investigated further. Another issue for future research is the origins of the (at least) two clusters that can be seen in Fig. 1 for the PBC. We suspect that these clusters originate when merging translations of the Bible of different coverage (e.g., translations of the Old and the New Testament versus translations of the New Testament only).

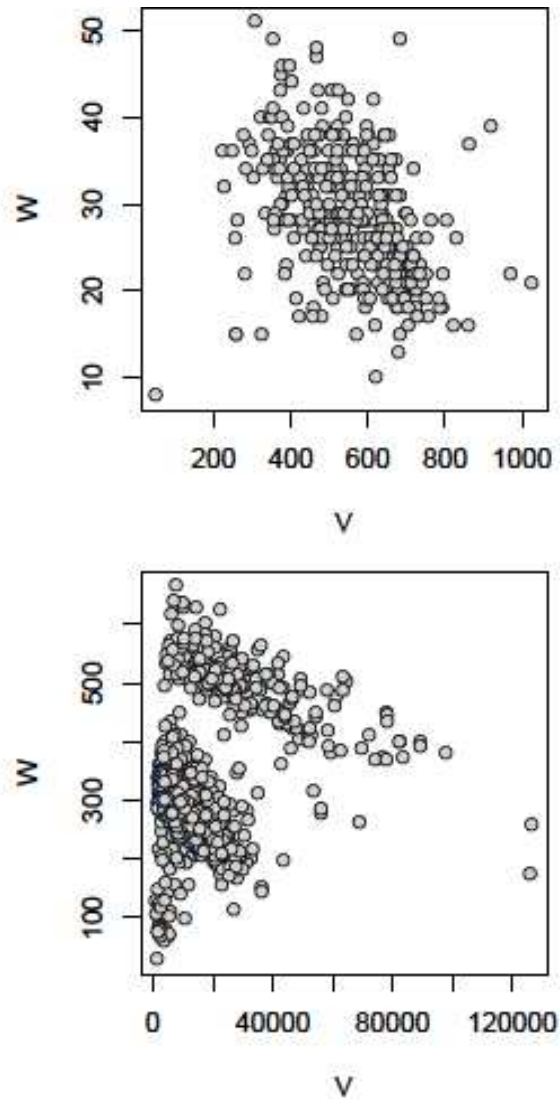

Figure 1. The number of different frequencies (*W*) versus the number of types (*V*) in parallel corpora. Every point corresponds to an individual text. Top: UDHR. Bottom: PBC.

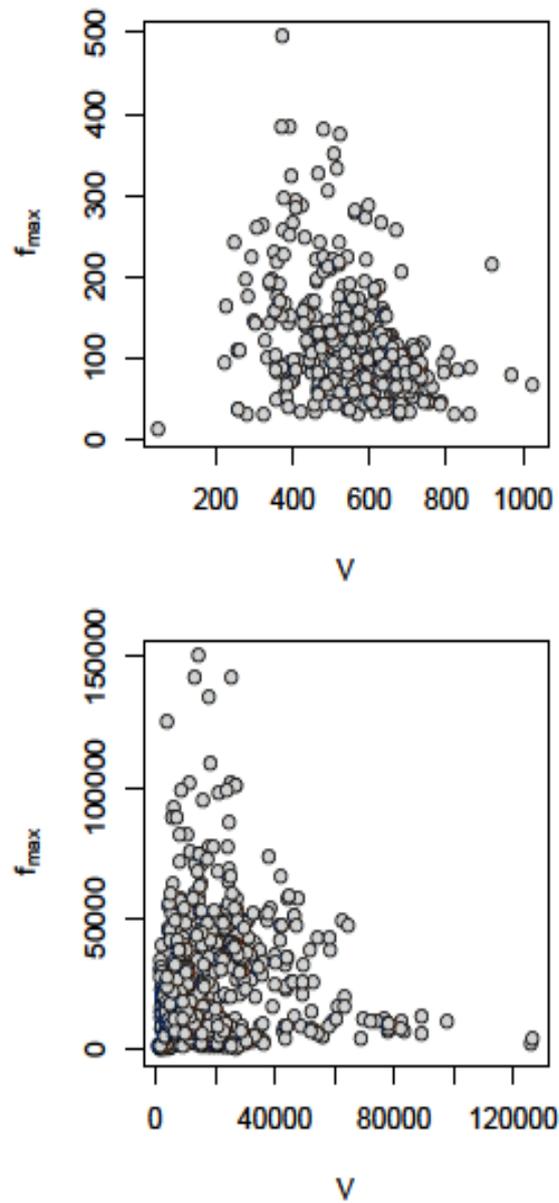

Figure 2. The maximum frequency ($f_{max}$) versus the number of types ($V$) in parallel corpora. Every point corresponds to an individual text. Top: UDHR. Bottom: PBC.

## Acknowledgements

BC and RFC are funded by the grant 2014SGR 890 (MACDA) from AGAUR (Generalitat de Catalunya). AL, BC and RFC are also funded by the APCOM project (TIN2014-57226-P) from MINECO (Ministerio de Economia y Competitividad). CB is funded by the DFG Center for Advanced Studies "Words, Bones, Genes, Tools" and the ERC grant EVOLAEMP at the University of Tübingen.